\documentclass[conference]{IEEEtran}
\IEEEoverridecommandlockouts
\usepackage{cite}
\usepackage{amsmath,amssymb,amsfonts}
\usepackage{graphicx}
\usepackage{textcomp}
\usepackage{xcolor}
\usepackage{listings}
\usepackage{algpseudocode,algorithm,algorithmicx}
\usepackage{tikz}
\usetikzlibrary{automata, positioning}
\usepackage{microtype}
\usepackage{booktabs}
\usepackage{pgfplots}
\usepackage{pgfplotstable}
\usepackage{caption} % Required for customizing captions
\usepackage{array}
\usepackage{hyperref}
\usepackage{subcaption}
\usepackage{comment}

\pgfplotsset{compat=1.16}

\def\BibTeX{{\rm B\kern-.05em{\sc i\kern-.025em b}\kern-.08em
T\kern-.1667em\lower.7ex\hbox{E}\kern-.125emX}}

\definecolor{keywordcolor}{rgb}{0.1, 0.1, 0.8}   % Blue keywords
\definecolor{commentcolor}{rgb}{0, 0.6, 0}       % Green comments
\definecolor{stringcolor}{rgb}{0.8, 0, 0.8}      % Purple strings

\begin{document}
\raggedbottom

\title{ICanC: Improving Camera-based Object Detection and Energy Consumption in Low-Illumination Environments\\
\thanks{This research was sponsored by the National Science Foundation Research Experience for Undergraduates Program.}
}

\author{\IEEEauthorblockN{Daniel Ma$^{\star}$, Ren Zhong$^{\dagger}$, and Weisong Shi$^{\ddagger}$}
\IEEEauthorblockA{$^{\star}$Khoury College of Computer Sciences, Northeastern University} 
\IEEEauthorblockA{$^{\dagger}$Department of Computer Science, Wayne State University} 
\IEEEauthorblockA{$^{\ddagger}$Department of Computer and Information Sciences, University of Delaware }
}

\maketitle 

\begin{abstract}
This paper introduces ICanC (\textit{pronounced "I Can See"}), a novel system designed to enhance object detection and optimize energy efficiency in autonomous vehicles (AVs) operating in low-illumination environments. By leveraging the complementary capabilities of LiDAR and camera sensors, ICanC improves detection accuracy under conditions where camera performance typically declines, while significantly reducing unnecessary headlight usage. This approach aligns with the broader objective of promoting sustainable transportation.

ICanC comprises three primary nodes: the \textit{Obstacle Detector}, which processes LiDAR point cloud data to fit bounding boxes onto detected objects and estimate their position, velocity, and orientation; the \textit{Danger Detector}, which evaluates potential threats using the information provided by the Obstacle Detector; and the \textit{Light Controller}, which dynamically activates headlights to enhance camera visibility solely when a threat is detected.

Experiments conducted in physical and simulated environments demonstrate ICanC's robust performance, even in the presence of significant noise interference. The system consistently achieves high accuracy in camera-based object detection when headlights are engaged, while significantly reducing overall headlight energy consumption. These results position ICanC as a promising advancement in autonomous vehicle research, achieving a balance between energy efficiency and reliable object detection.

\end{abstract}


\section{Introduction}
In low-illumination environments, vehicle headlights are essential for ensuring safety and visibility, especially for drivers and passengers. Autonomous vehicles (AVs), which rely heavily on camera sensors for navigation, also need headlights to improve visibility and provide clearer images for decision-making. However, headlights are not always necessary for safe operation, particularly when utilizing versatile sensors such as LiDAR, which can offer consistent performance regardless of lighting conditions. Minimizing headlight usage in AVs presents a promising strategy for reducing energy consumption without compromising safety or detection capabilities.

The most prevalent approach to AV perception typically involves camera sensors, which capture 2D images of the environment. These images provide detailed visual information but lack depth perception and are heavily dependent on lighting conditions. This reliance on cameras can become problematic in low-light or adverse weather conditions. For instance, in 2019, a Tesla vehicle operating in autopilot mode collided with a truck before sunrise, leading to a fatal accident. The incident was attributed to the Autopilot system's inability to detect the truck in time due to poor visibility in low-light conditions. As Tesla’s system relied solely on cameras, it failed to “consistently detect and track the truck as an object or threat as it crossed the path of the car\cite{thadani_2023_the}." This tragic incident underscores the limitations of camera-based perception in low-illumination environments and highlights the need for more robust detection systems in these conditions.

\begin{table}
    \centering
    \begin{tabular}{lccc}
        \toprule
        & \textbf{Day} & \textbf{Night} & \textbf{Night w/Flashlight} \\
        \midrule
        LiDAR Detection Rate (\%) & 100 & 87.5 & 100 \\
        Camera Detection Rate (\%) & 100 & 25 & 100 \\
        \bottomrule
    \end{tabular}
    \caption{Percent of pedestrians detected by LiDAR and camera algorithms, as recorded during experimental observations under varying conditions.}
    \label{tab:detection_rates}
\end{table}

To overcome these limitations, the combination of multiple sensors—specifically cameras and LiDAR—holds significant promise. LiDAR, or light detection and ranging, emits laser beams in a defined field of view, often in a 360-degree cycle, and measures the time it takes for the beams to return. This process generates a 3D point cloud, where each point represents coordinates and additional attributes. Due to its 3D nature, this point cloud contains depth information that camera images lack. Furthermore, these lasers operate independently of the illumination state, allowing the advantage of nighttime usage. This trait is evident from experimental observations, which show that LiDAR detection rates only degrade approximately 12.5\% in the dark, while camera detection rates degrade by 75\% (see Table \ref{tab:detection_rates}). However, LiDAR has its own limitations as well. Firstly, lasers are similarly prone to noise, or unwanted and irrelevant variations in data, from adverse weather conditions. Additionally, while LiDAR sensor costs have decreased over time, they remain more expensive than camera sensors\cite{zamanakos_2021_a}. Despite these challenges, the fusion of camera and LiDAR data allows for the strengths of each sensor to compensate for the other's weaknesses, making it an ideal solution for enhancing object detection in low-light environments.

By fusing the data from both cameras and LiDAR, headlight usage may be optimized while maintaining high detection performance in the dark. When no danger is detected, the headlights can remain off, saving energy. When a potential threat is identified, the headlights can be activated to enhance camera performance. This selective use of headlights reduces unnecessary energy consumption without sacrificing safety or detection reliability.

To implement this concept, this project introduces ICanC—a novel system designed to optimize headlight usage while ensuring effective object detection in low-light environments through sensor fusion. ICanC consists of three primary components: Firstly, the \textit{Obstacle Detector} node, which processes LiDAR point clouds to identify objects and tightly fit 3D bounding boxes around them. This data is passed to the \textit{Danger Detector} node, which analyzes bounding boxes to identify potential threats in each frame. Analysis of data is then passed to the \textit{Light Controller} node, which activates headlights when danger is detected, enabling the camera to capture clearer images of the surroundings. Through sensor fusion of LiDAR and camera data, ICanC ensures that detection performance remains strong in low-light conditions, while headlight usage is minimized to optimize energy consumption. This approach not only improves safety in autonomous driving but also supports the broader goal of sustainable transportation by reducing the environmental impact of electric autonomous vehicles.

%new paragraph
Furthermore, this project aims to demonstrate the Vehicle Programming Interface (VPI) and its potential to support the efficiency of programming with autonomous vehicles. The VPI was developed as a response to the growing interdisciplinary complexities of automotive systems, which make developing applications for AVs more challenging \cite{wu_2024_vpi}. Much of the logic implemented in ICanC can be abstracted into the VPI and employed to solve various problems, thus promoting the goal of accessibility of AV programming.

In conclusion, this paper makes the following contributions:
\begin{enumerate}
    \item Defines the low-illumination environment problem for object detection in AVs.
    \item Introduces ICanC, a novel system that improves low-illumination object detection and simultaneously reduces energy consumption for AVs. ICanC is open-source \footnote{The code for ICanC is available at \href{https://github.com/danielma4/ICanC_v2}{https://github.com/danielma4/ICanC\_v2.}}.
    \item Evaluates the system in both extensive real-world and simulation tests.
\end{enumerate}

This paper begins by reviewing related literature in Section \hyperref[sec:2]{II} before presenting an in-depth explanation of the ICanC design in Section \hyperref[sec:3]{III}. Section \hyperref[sec:4]{IV} then discusses the implementation of ICanC in physical and simulated environments. Section \hyperref[sec:5]{V} showcases results gathered from extensive experimentation and testing, and Section \hyperref[sec:6]{VI} then discusses these results, potential improvements, and future works. Section \hyperref[sec:7]{VII} then summarizes and concludes the project. Finally, Section \hyperref[sec:8]{VIII} offers acknowledgments to those who provided invaluable support during the project.

\section{Related Works}
\label{sec:2}
Related works can be classified into two research areas: improving low-illumination object detection and energy conservation in autonomous vehicles.

\subsection{Low-illumination object detection}
Current low-illumination AV research utilizes a multitude of techniques to improve performance of detection algorithms. One such approach is \textit{sensor fusion}, or using multiple sensors in conjunction. In particular, synergizing camera images with thermal imaging has shown success in surveillance object detection. Thermal sensors record changes in the temperature of an environment, allowing for an image largely unaffected by the absence of light. By combining images outputted by cameras and thermal imaging, the best of both sensors can be meshed into an improved snapshot\cite{patel_2020_night}. Another sensor fusion technique is gated imaging. Laser-gated imaging operates by emitting laser photons in tandem with a synchronized imaging sensor. This photon offers a brief source of light and illuminates the area. The reflected light returns to the imaging sensor, offering information about the surroundings. Research has experimented with gated imaging in autonomous vehicles, providing advantages of 2D and 3D object detection and performing well in both low-illumination and adverse weather conditions\cite{julcaaguilar_2021_gated3d}. While ICanC builds on sensor fusion delineated in the above research, this system explores the potential of LiDAR rather than thermal or gated imaging. Furthermore, the ICanC system employs more straightforward logic by keeping the imaging of both sensors separate rather than combining the two into a fused image or sensor. This distinction allows ICanC to leverage the unique strengths of each sensor independently, offering a simpler yet effective approach to enhancing object detection in low-illumination environments.

Another approach is \textit{image preprocessing}. This technique can be accomplished through numerous techniques, including artificially brightening images and extracting pertinent information. Research utilizing Gaussian Process regression has shown success in retrieving relevant information rather than aesthetic restoration; however, artificially altering images inevitably introduces noise and can prove inaccurate at times \cite{loh_2019_lowlight}. Furthermore, models trained on altered low-illumination images have not shown improved results over models only trained on low-illumination images \cite{xiao_2020_making}. While image preprocessing carries with it the potential to drastically enhance low-illumination object detection if done in real-time, this project has elected to focus on the capabilities of sensor fusion.

\subsection{Improving energy conservation in autonomous vehicles}
Research on optimizing energy efficiency in AVs has been diverse and effective, with many projects focusing on various components relevant to autonomous vehicles and detection algorithms. These works include applying the energy-efficient Spiking Neural Networks (SNN) to camera-based object detection, optimizing backbone networks for real-time detection algorithms, and assigning suboptimal, yet highly efficient detection algorithms to subsets of sensors in camera networks \cite{kim_2020_spikingyolo, lee_2019_an, dao_2017_energy}; however, to the best of the authors' knowledge, no work has actively taken advantage of low-illumination to lower energy consumption in AVs. Autonomous vehicle research is prominent, and AVs will continue to improve in these areas.

\section{ICanC Design}
\label{sec:3}
\begin{figure}
    \centering
    \includegraphics[width=1\linewidth]{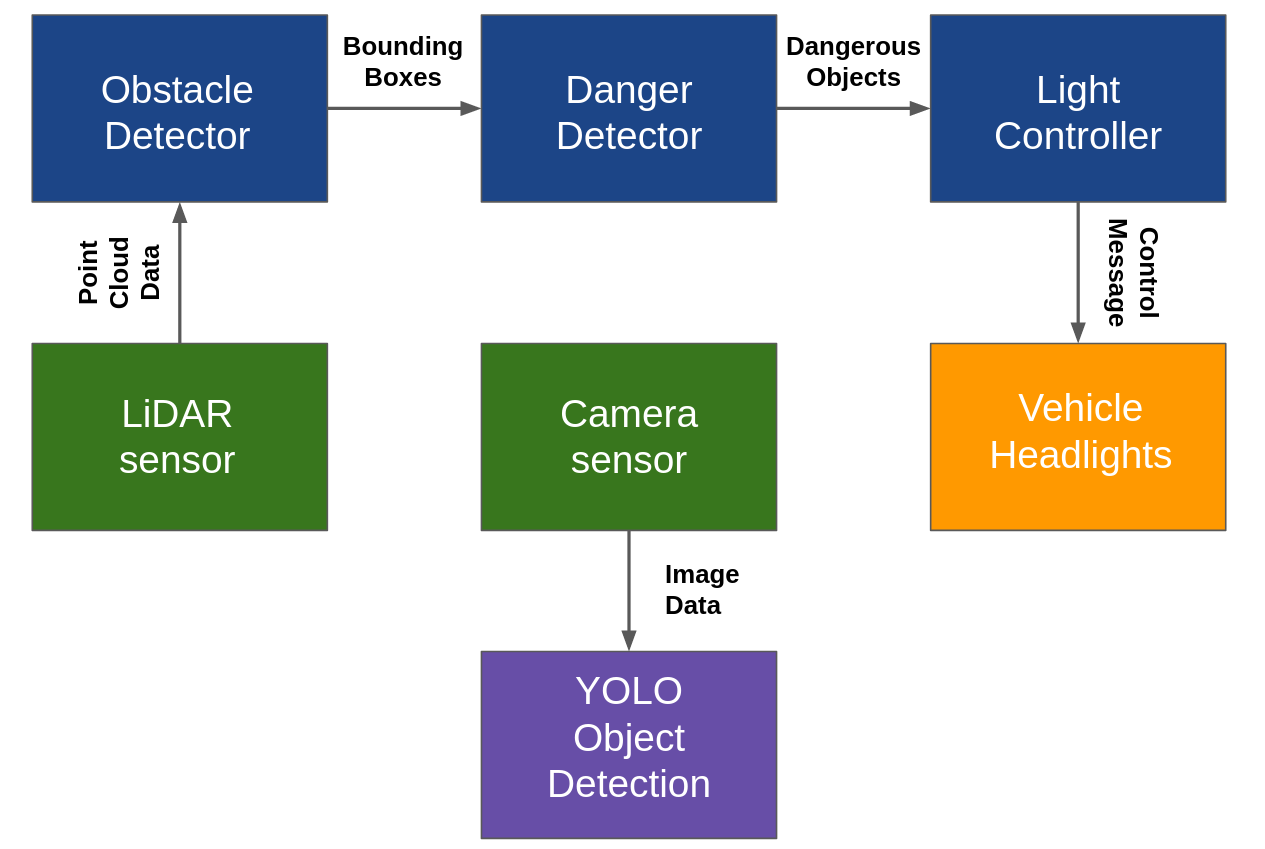}
    \caption{ICanC system overview}
    \label{fig:1}
\end{figure}
\subsection{Overview}
The ICanC system is designed to enhance vehicle safety by detecting obstacles, assessing potential dangers, and controlling vehicle lights to improve visibility. The system is composed of three interconnected nodes: the Obstacle Detector node, the Danger Detector node, and the Light Controller node. Each node performs specific functions that contribute to the overall functionality of the ICanC system.

The system overview in Fig. \ref{fig:1} illustrates the interactions and relationships between these nodes, providing a visual representation of the system's operation. To utilize the system, only sensor data and control over headlights need to be given to ICanC; this promotes transferability and in theory, allows usage of ICanC across numerous platforms.

\subsection{Methodology}
The ICanC system comprises three nodes. Firstly, LiDAR sensors output point cloud data to the \textbf{Obstacle Detector} node, whose primary purpose is to identify objects and tightly fit 3D bounding boxes to each object. Next, these bounding boxes are sent to the \textbf{Danger Detector} node, which utilizes a variety of available data to determine whether or not each detected object has potential for danger to the vehicle. These dangerous objects are then collected and sent to the \textbf{Light Controller} node, which simply controls the light, allowing the camera sensors better imaging in low-illumination. These nodes are each discussed in full detail in the following subsections.

\subsubsection{Obstacle Detector Node}
There are numerous LiDAR object detection algorithms available for potential use in autonomous vehicles. The algorithm primarily used in the Obstacle Detector is Euclidean Clustering. It works as follows: the LiDAR sensors will formulate a point cloud frame, sending this data to this node. The node will iterate over each point in this cloud, determining the Euclidean distance between other points in the cloud and clustering points within a set radius of the current point. Through this technique, a cluster of points returned from the sensors is grouped, allowing the node to tightly fit a box around the cluster's edge points.

Additionally, this node calculates information relevant to the Danger Detector node. The Obstacle Detector keeps track of consecutive frames. With information concerning bounding boxes calculated, this node compares these consecutive frames and determines which objects are the same between frames. It does so by comparing overall displacement and intersection over union (IOU) between potential matches, collecting potential matches in a list, and then applying the Hungarian matching algorithm, assigning optimal matches to objects between frames.

With information from matching boxes identified, more important attributes relevant to danger detection can be calculated. Given two boxes representing the same object through consecutive frames, velocity can be determined in m/s as a calculation of translation between boxes divided by the time between the creation of the boxes. It should be noted that the node takes the translation of the vehicle as well as the translation of the object into consideration; however, the LiDAR sensor on the vehicle is the fixed frame, so the vehicle's translation is simply applied to the object's translation as well. Next, the orientation of the object can be calculated in the current frame through the angle between vectors formula:
\[
\theta = \cos^{-1}\bigg(\frac{\vec{u} \cdot \vec{v}}{||\vec{u}||\cdot||\vec{v}||} \bigg),
\]
where u and v are the x and y coordinates of the boxes $\in \mathbb{R}^{2}$ after translating the positions of the objects to the origin\footnote{Only yaw, or rotation about the z-axis, is considered in orientation calculation.}.

Early iterations of the ICanC system suffered from noisy and inaccurate sensor readings. Positions of bounding boxes were seemingly spontaneously assigned, leading to erroneous calculations. To combat faulty and inconsistent readings, a Kalman filter was implemented in this node to estimate the states of object positions and velocities in the 3-dimensional plane. In this filter, velocity is assumed to be constant. Previously calculated velocity and positions are fed to the filter, and readings are updated accordingly for orientation as well. Through filtering observed readings across frames, measurements are smoothed out, effectively combating unpredictable and noisy sensor data.

With all the necessary data calculated and filtered, an array of bounding boxes is sent to the Danger Detector node, each box carrying information on its position, velocity, and orientation.

\subsubsection{Danger Detector Node}
\begin{figure}
    \centering
    \begin{tikzpicture}[scale=0.9] % Adjust the scale factor here
        % Shading the regions x > 3 and x < -3
        \fill[orange!20] (0,0) -- (-5,5) -- (-5,0) -- cycle;
        \fill[red!20] (0,0) -- (-5,5) -- (5,5) -- cycle;
        \fill[green!20] (0,0) -- (5,5) -- (5,0) -- cycle;

        % Drawing the axes
        % Draw y-axis
        \draw[->] (0,0) -- (0,5) node[anchor=south] {$x$};
        % Draw x-axis
        \draw[->] (0,0) -- (-5,0) node[anchor=east, shift={(0.5,-0.3)}] {$y$};
        \draw[->] (0,0) -- (5,0);

        % Adding labels for shaded regions
        \node at (4, 2) {Section 3};
        \node at (1.75, 4) {Section 2};
        \node at (-4, 2) {Section 1};

        % Adding the boundaries
        \draw[dashed] (0,0) -- (5,5);
        \draw[dashed] (-5,5) -- (0,0);
        
        % Adding object in Section 1
        \filldraw[blue] (-2.5, 1) rectangle (-2, 1.5);
        % Adding FOV
        \draw[dashed] (-2, 1.25) -- ++(90:1.3);
        \draw[dashed] (-2, 1.25) -- ++(270:1.1);
        \draw (-2, 1.09) arc[start angle=-90, end angle=90, radius=.15];

        \filldraw[blue] (2, .75) rectangle (2.5, 1.25);
        % Adding FOV
        \draw[dashed] (2, 1) -- ++(90:1.3);
        \draw[dashed] (2, 1) -- ++(270:1);
        \draw (2, 1.155) arc[start angle=90, end angle=270, radius=.15];

        \filldraw[blue] (-1.25, 4) rectangle (-.75, 3.5);
        \draw (-.555, 3.75) arc[start angle=0, end angle=360, radius=.45];

        \node[font=\scriptsize] at (-1.8, 3.75) {$360^\circ$};
        \node[font=\scriptsize] at (1.5, 1) {$180^\circ$};

        \draw (-.2, .2) arc[start angle=135, end angle=180, radius=.3];
        \node[font=\scriptsize] at (-.5, .2) {$45^\circ$};

        \begin{scope}[rotate=45]
            \draw (0,0) rectangle (0.28,0.28);
        \end{scope}
    \end{tikzpicture}
    \caption{Initial Danger Detector design}
    \label{fig:old_design}
\end{figure}
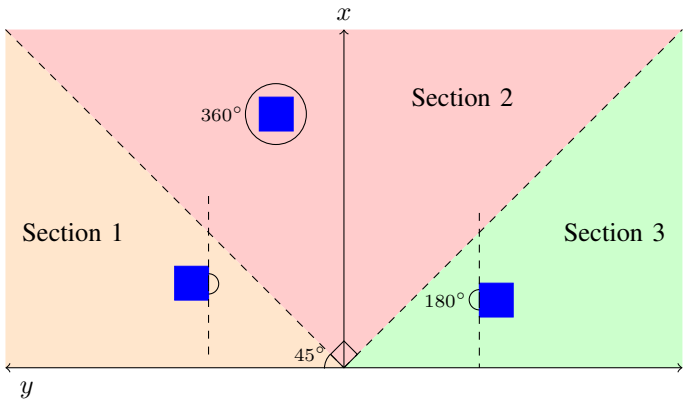

The initial design of the Danger Detector is shown in Fig. \ref{fig:old_design} and is as follows. The Danger Detector splits the frontal 180$^\circ$ field-of-view (FOV) of the vehicle into three distinct sections: the leftmost $45^\circ$, the center $90^{\circ}$, and the rightmost $45^\circ$. The logic for the danger detector relies on mainly two factors: velocity and orientation. However, how these two are handled differently in the side and center sections.

\textbf{Side sections}: If an object is determined to be present in the side FOV of the vehicle, the orientation of the object is taken heavily into account. Specifically, if the object's previously calculated yaw faces the $180^\circ$ toward the vehicle, the Danger Detector determines that the object is facing the vehicle. More specifically, if an object to the left of the vehicle is facing anywhere from $180^\circ$ to $360^\circ$\footnote{Yaw orientation starts with $0^\circ$ facing directly ahead of the vehicle, and $180^\circ$ is facing directly behind the vehicle.}, the Danger Detector deems it facing the vehicle. For the rightmost FOV section, if it faces anywhere from $0^\circ$ to $180^\circ$, it is facing the vehicle. 

Next, the speed is taken into account. Given a float parameter $t$ representing reaction time allowed, the distance at which the node deems a particular object dangerous is calculated. The system aims to allow for $t$ seconds for the vehicle or passenger to react to the object after the light has been turned on. So, the distance formula $d = s \times t$ is used to calculate the approximate distance the current object will travel within $t$ seconds. 

Bringing these two calculations together, if the object's distance from the vehicle is currently less than the distance which the object will travel in $t$ seconds and the object has been deemed facing the vehicle, the object has the potential to cause danger and the node labels the current object dangerous.

\textbf{Front section}: The danger of an object directly in the path of the vehicle is a more immediate issue. So in this case, the node disregards orientation altogether. Otherwise, the logic remains the same. If the distance of the object to the vehicle is less than the distance which the object will travel within $t$ seconds, the Danger Detector deems the object dangerous.

There are a few flaws evident in this initial design, each contributing to the overall issue in the implementation: the danger label was applied indiscriminately to objects, outputting numerous false positives. Firstly, the center section covered a vast area, equal to half of the vehicle's FOV. Considering that the Danger Detector was considerably more lenient with objects found in this section, ignoring the orientation of the object altogether, multiple detected objects were given a dangerous label. To put this into perspective, Fig. \ref{fig:old_design} demonstrates the range of the center section. Specifically, an object directly ahead of the vehicle could be given the same level of danger as an object far to the left or right of the vehicle. In the side sections, where orientation is taken into account, the $180^\circ$ FOV toward the vehicle allowed for objects was also overly lenient. In particular, an object could be moving parallel to the vehicle, never to collide, yet the Danger Detector would place the danger label on the object indefinitely. Overall, the initial design contained a few flaws, causing the system to designate far too many objects as dangerous.
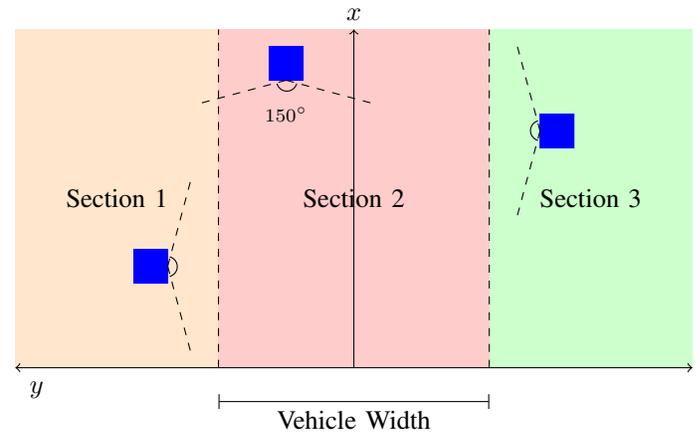
\begin{figure}
    \centering
\begin{tikzpicture} [scale=.9]
    % Shading the regions x > 3 and x < -3
    \fill[green!20] (2,0) rectangle (5,5);
    \fill[orange!20] (-5,0) rectangle (-2,5);
    \fill[red!20] (-2, 0) rectangle (2, 5);

    % Drawing the axes
    % Draw y-axis
    \draw[->] (0,0) -- (0,5) node[anchor=south] {$x$};
    % Draw x-axis
    \draw[->] (0,0) -- (-5,0) node[anchor=east, shift={(0.5,-0.3)}] {$y$};
    \draw[->] (0,0) -- (5,0);

    % Adding labels for shaded regions
    \node at (3.5, 2.5) {Section 3};
    \node at (0, 2.5) {Section 2};
    \node at (-3.5, 2.5) {Section 1};

    % Adding the boundaries
    \draw[dashed] (-2,0) -- (-2,5);
    \draw[dashed] (2,0) -- (2,5);

    % Line with ticks and label
    \draw[|-|] (-2, -.5) -- (2, -.5) node[midway, below] {Vehicle Width};

    % Adding object in Section 1
    \filldraw[blue] (-3.25, 1.25) rectangle (-2.75, 1.75);
    % Adding FOV
    \draw[dashed] (-2.75, 1.5) -- ++(75:1.3);
    \draw[dashed] (-2.75, 1.5) -- ++(285:1.3);
    \draw (-2.72, 1.35) arc[start angle=-75, end angle=75, radius=.15];

    \filldraw[blue] (2.75, 3.25) rectangle (3.25, 3.75);
    % Adding FOV
    \draw[dashed] (2.75, 3.5) -- ++(105:1.3);
    \draw[dashed] (2.75, 3.5) -- ++(255:1.3);
    \draw (2.72, 3.65) arc[start angle=105, end angle=255, radius=.15];

    \filldraw[blue] (-1.25, 4.75) rectangle (-.75, 4.25);
    % Adding FOV
    \draw[dashed] (-1, 4.25) -- ++(195:1.3);
    \draw[dashed] (-1, 4.25) -- ++(345:1.3);
    \draw (-1.135, 4.2) arc[start angle=195, end angle=345, radius=.15];

    \node[font=\scriptsize] at (-1, 3.75) {$150^\circ$};

    \end{tikzpicture}
        \caption{Current Danger Detector design}
    \label{fig:new_design}
\end{figure}

The current design for the Danger Detector node is shown in Fig. \ref{fig:new_design} and improves upon these issues in a few ways. Firstly, the three sections are defined differently. Given a float parameter $w$ representing the width of the vehicle, the Danger Detector splits the vehicle's frontal FOV into three sections: the direct path from $-w < y < w$, the leftmost FOV $y \ge w$, and the rightmost FOV $y \le -w$. This adjustment decreases the overall area of the more dangerous direct path of the vehicle. Behavior for each section is defined below.

\textbf{Side sections}: The orientation allowed for objects detected in the side sections has been decreased from 180$^\circ$ to $150^\circ$. More specifically, for objects in the left FOV of the vehicle, if the Obstacle Detector node calculated the object's orientation to be between 195$^\circ$ to 345$^\circ$, the Danger Detector labels it as facing the vehicle. For objects in the right section, the FOV allowed for danger is from 15$^\circ$ to 165$^\circ$ rather than 0$^\circ$ to $180^\circ$. This increases the criteria for dangerous objects and eliminates the issue of objects moving parallel to the vehicle being deemed dangerous. Otherwise, the logic remains the same.

\textbf{Front section}: Previously, the Danger Detector disregarded orientation altogether with the idea that an object within the direct path of the vehicle would be dangerous regardless of orientation. However, considering that the translation calculation from the Obstacle Detector takes into account the vehicle's movement and simply applies the vehicle's translation to the object, the orientation of the object should factor into the logic. For example, if an object had orientation 0$^\circ$, or away from the vehicle, that means that the overall displacement of the object is away from the vehicle from one frame to the next. Since the object is moving away, it shouldn't be deemed dangerous. To combat this, logic from the side sections is implemented into the front section. The Danger Detector allows for $150^\circ$ FOV toward the vehicle to be deemed facing the vehicle, from $105^\circ$ to $255^\circ$. Other than the addition of orientation into the logic, the overall calculation remains the same for determining if an object is dangerous.

The Danger Detector collects all objects deemed dangerous into an array of bounding boxes and passes them into the Light Controller node.

\subsubsection{Light Controller Node}

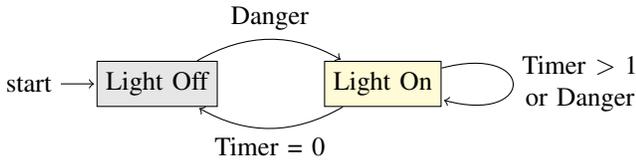
\begin{figure}
\centering
\begin{tikzpicture}[shorten >=1pt, node distance=3cm, on grid, auto]

   % Define the style for the states
   \tikzstyle{state}=[rectangle, draw, fill=gray!20, minimum size=6mm]
   
   % Define the states
   \node[state, initial] (off) {Light Off};
   \node[state, fill=yellow!20] (on) [right=of off] {Light On};
   
   % Define the paths
   \path[->]
    (off) edge[bend left, above] node{Danger} (on)
    (on) edge[bend left, below] node{Timer = 0} (off)
    (on) edge[loop right] node[align=center]{Timer $>$ 1 \\ or Danger} (on);
    
\end{tikzpicture}
\caption{State diagram for the Light Controller}
\label{fig:light_controller}
\end{figure}
The Light Controller has two states, shown in Fig. \ref{fig:light_controller}, in which it can reside: light on or off. Each of these states carries with it differing logic. When this node receives data from the Danger Detector in the off state, the Light Controller turns the light on. Given an integer parameter $\tau$, the Light Controller will also start a timer for $\tau$ seconds, during which the light will continue to shine regardless of the presence or absence of danger. This time frame helps to eliminate any inconveniences that could arise from the light flashing on and off frequently. While the Light Controller is in the light-on state, it will ignore any new messages until the timer reaches one second. When the timer reaches one second, the light will still be on, but the Light Controller will begin to check whether or not the LiDAR still detects danger. If the Danger Detector passes new information to the Light Controller during this time frame, the timer will reset and the light will continue to shine for a new cycle. If no new information is received from the Danger Detector, the Light Controller will turn the light off and the timer will stay at zero. From here, the logic cycle repeats.

\section{Implementation}
\label{sec:4}
This system is built using the open-source Robot Operating System (ROS). Each node mentioned in the previous section is a ROS node, equipped with publishers and subscribers to communicate with one another through ROS topics. The LiDAR and camera object detection algorithms utilized were written with ROS wrappers, allowing for compatibility with the system. The ICanC system employs a Euclidean clustering LiDAR object detection algorithm as well as the single-pass YOLOv3 camera detection algorithm \cite{sun_2021_lidar, redmon_2018_yolov3, bjelonic_2016_yolo}. The Obstacle Detector also employs a C++ implementation of a Kalman filter \cite{pei_2019_an, martiros_2014_hmartirokalmancpp}. 

Concerning the functions written for ICanC, the code and logic outlined in Section \hyperref[sec:3]{3} can be abstracted and implemented into the Vehicle Programming Interface (VPI) to address a broader range of challenges in autonomous driving. Three key functions have been identified for their potential versatility: \textit{getSection}, which determines whether an object is to the left, middle, or right of the vehicle; \textit{isMovingTowardMe}, which assesses whether a detected object is moving toward the vehicle; and \textit{getSpeedOfObject}, which calculates an object's speed. The pseudocode for ICanC's danger detecting logic implemented with the VPI is provided in Listing \ref{lst:danger}. This system is one such case where these functions prove useful, but these functions also may play critical roles in applications such as obstacle detection, collision avoidance, adaptive cruise control, and pedestrian tracking. By generalizing these functions into the VPI, both the efficiency and accessibility of autonomous vehicle programming is enhanced.\\

\begin{figure}
    \centering
    \includegraphics[width=1\linewidth]{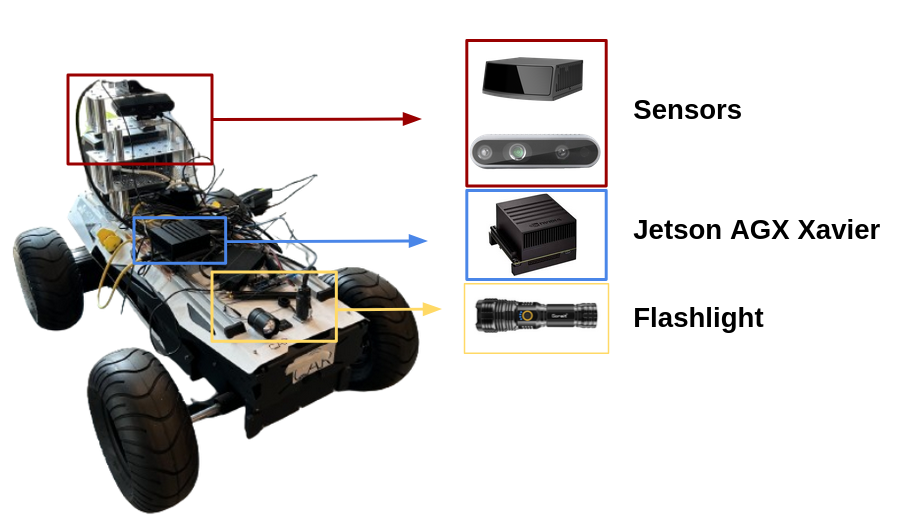}
    \caption{Zebra: A general use indoor robot}
    \label{fig:zebra}
\end{figure}

\begin{lstlisting}[caption={Danger Detecting with VPI},label={lst:danger}]
import vpi

detect_danger(boxes):
    reaction_time_allowed = 3 #in seconds
    
    for box in boxes:
        speed = vpi.getSpeedOfObject(box)
        section = vpi.getSection(box)
        dist = norm(box) #dist from vehicle to object in meters

        danger_threshold = speed * reaction_time_allowed

        if (dist <= danger_threshold and
        vpi.isMovingTowardMe(box, section)):
            dangerous_boxes.append(box)

    return dangerous_boxes
\end{lstlisting}

This system was evaluated in three testing environments: a physical environment, a simulation environment via the CARLA autonomous driving simulator, and a Monte Carlo simulation.

\subsection{Physical Testing}
The ICanC system was installed onto the CAR Lab's Zebra, a general-purpose indoor autonomous vehicle, and demonstrated in various lighting conditions and environments.

\subsubsection{Zebra}
The Zebra's computing unit is the NVIDIA Jetson AGX Xavier Developer Kit running Ubuntu 18.04 and ROS Melodic \cite{nvidia}. This autonomous vehicle is shown in Fig. \ref{fig:zebra}, equipped with a Robosense LiDAR M1 sensor as well as an Intel RealSense DepthCamera D435i, which features an RGB sensor with maximum resolution of 1920$\times$1080 \cite{robosense, intel}. Furthermore, the Zebra's chassis is the AgileX hunter robot, which integrates an Ackermann control-based drive-by-wire \cite{a2024_hunter}. This allows for both manual and autonomous control of the vehicle. To simulate headlights for the Zebra, a Goreit flashlight was interfaced with Arduino and attached to the vehicle\cite{a2014_amazoncom, arduino}.

%start here
\subsubsection{Course/Setup}
\begin{figure}
\begin{subfigure}{0.5\textwidth}
\includegraphics[width=1\linewidth, height=5cm]{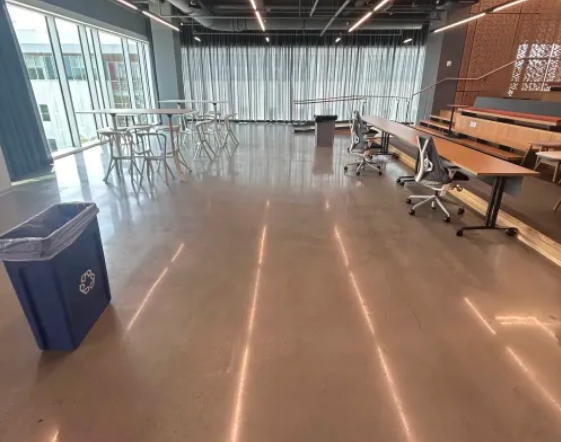} 
\caption{One physical environment used in testing.}
\label{fig:subim1}
\end{subfigure}
\\

\begin{subfigure}{0.5\textwidth}
\includegraphics[width=1\linewidth, height=5cm]{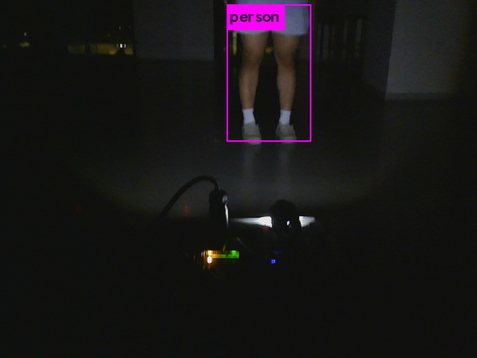}
\caption{Camera object detection in physical environment}
\label{fig:subim2}
\end{subfigure}
\caption{Physical testing}
\label{fig:physical}
\end{figure}
Testing was conducted in the FinTech building at the University of Delaware at multiple sites. One such course is shown in Fig. \ref{fig:subim1} and \ref{fig:subim2}, arbitrarily populated with objects, such as garbage bins, chairs, and tables, to challenge the ICanC system. The testing took place at nighttime, with building lights on and off. The tests were also run with and without the flashlight attached to the Zebra to highlight the differing capabilities of LiDAR and camera sensors in the dark. Furthermore, a pedestrian was introduced to the scene, moving toward, away from, and perpendicular to the Zebra in different tests.

\subsection{Simulation Testing}
The CARLA simulator provides realistic environments with control over the time of day, sensors, streetlights, and headlights, allowing accurate simulation of environments where the ICanC system would be effective \cite{dosovitskiy_2017_carla}.

\subsubsection{Vehicle}
The vehicle equipped with ICanC in the simulation is the Tesla Model 3. This vehicle is armed with a custom LiDAR sensor with 128 channels and 640,000 points per second generated, as well as an RGB camera sensor. Parameters were adjusted to accommodate the Tesla's width $w$ and overall FOV of the LiDAR to allow for a wider radius of detection than the smaller Zebra. Fig. \ref{fig:subim3} and \ref{fig:subim4} shows the vehicle in the pitch-black environment, as well as its LiDAR and camera detection outputs in Fig. \ref{fig:subim5} and \ref{fig:subim6}. This vehicle was equipped with both autonomous and manual control capabilities, and both were employed to observe the behavior of the system.

\subsubsection{Environment}
\begin{figure}
\begin{subfigure}{.5\textwidth}
\includegraphics[width=1\linewidth, height=5cm]{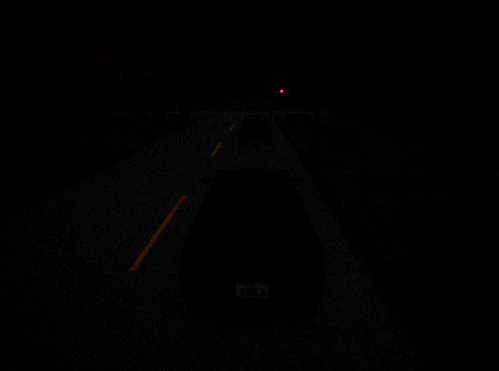} 
\caption{ICanC equipped vehicle approaching another vehicle in the dark.}
\label{fig:subim3}
\end{subfigure}
\\

\begin{subfigure}{0.5\textwidth}
\includegraphics[width=1\linewidth, height=5cm]{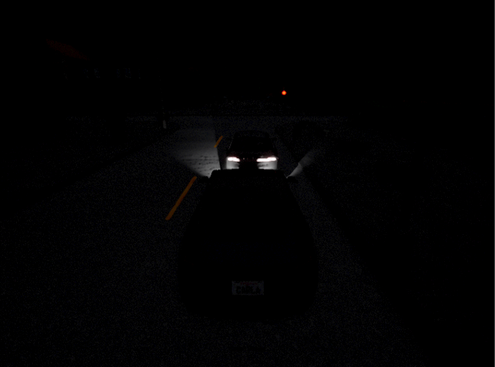}
\caption{Other vehicle deemed as dangerous, headlights activate}
\label{fig:subim4}
\end{subfigure}
\\

\begin{subfigure}{0.5\textwidth}
\includegraphics[width=1\linewidth, height=5cm]{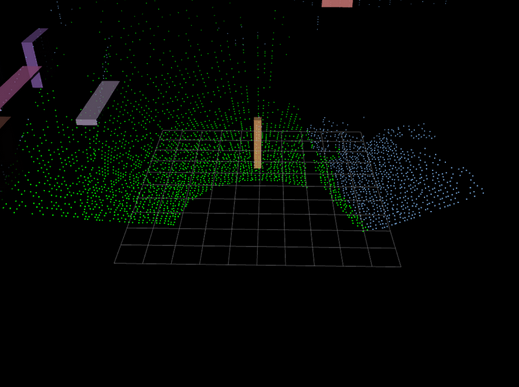}
\caption{Obstacle Detector detects an object directly ahead of sensors.}
\label{fig:subim5}
\end{subfigure}
\\

\begin{subfigure}{0.5\textwidth}
\includegraphics[width=1\linewidth, height=5cm]{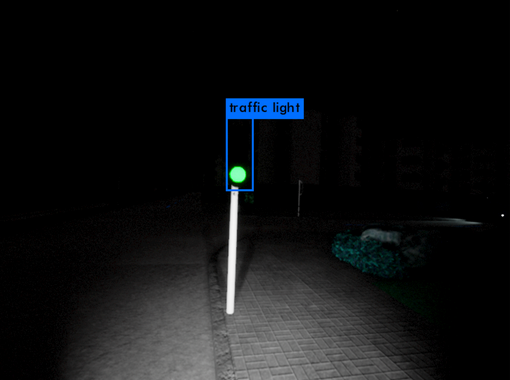}
\caption{Object is deemed dangerous, camera object detection identifies it as a traffic light.}
\label{fig:subim6}
\end{subfigure}
\caption{CARLA simulator testing}
\label{fig:carla}
\end{figure}
Testing was done in various preloaded environments offered by CARLA, including Town01. To achieve the desired world/environmental state, scripts were run to change the time of day to midnight, set all streetlights to off, and spawn numerous cars, trucks, and pedestrians in random areas. This created a pitch-black environment, as shown in Fig. \ref{fig:carla}. 

From there, ICanC was activated, allowing point cloud and camera data to be fed into the system's nodes.

\subsection{Monte Carlo Simulation}
The goals of this Monte Carlo simulation are to:
\begin{itemize}
    \item Determine the relationship between the number of objects detected in a frame and the percent of those objects deemed dangerous.
    \item Determine the average percentage of objects that the Danger Detector deems dangerous.
    \item Utilize this information to calculate approximate energy conservation.
\end{itemize}
The hypothesis before running the simulation was that the distribution would be uniform and that there was no special relationship between load size, or number of objects detected, and percent danger detected. 

To support this hypothesis, the probability of danger was calculated beforehand. Randomly generated bounding boxes representing detected objects were created with the following random variables: distance $d$ pulled from a discrete uniform distribution unif\{1, n\} in meters, speed $s$ pulled from the discrete uniform distribution unif\{1, n\} in meters per second, and orientation $\psi$ pulled from a discrete uniform distribution unif\{1, 360\} in degrees. The probability of a detected object being deemed dangerous is as follows.

Let $\theta$ be the danger distance threshold, defined as 
\[
\theta = \alpha \cdot s,
\]
where $\alpha$ represents the parameter for reaction time allowed for the vehicle. In this simulation, $\alpha = 1$, leaving us with $\theta = s$. Let $\textit{E}_{1}$ be the event that $d \le \theta$, or that the current position of the detected object is less than the danger threshold calculated in $\theta$, and let $\textit{E}_{2}$ be the event that the detected object is facing the vehicle. The event space $\Omega$ is defined through the two events and their four total combinations. 
Finally, let $D$ be the event that the detected object is dangerous.

Assuming that $\textit{E}_{1}$ and $\textit{E}_{2}$ are independent variables, the probability of \textit{D} can be defined as
\[
    P(D) = P(E_{1}) \land P(E_{2})
\]

In the above equation, \textit{P$(E_{2}$)} can be defined as $\frac{150}{360} \approx .4166$, as this is the allowed FOV to be deemed facing the vehicle. 

For $P(E_1)$, defined through the cumulative distribution function $P(E_1) = P(x \le \theta)$, the probability of each discrete point from 1 to $n$ needs to be considered. Let event $C_i$ be the event that the object is $i$ meters away from the vehicle, where $i \in [1, n]$. $P(E_1)$ can be expanded using the law of total probability.
\begin{align*}
P(E_1) &= \sum_{i = 1}^{n}P(E_{1}|C_i)\cdot P(C_i)\\
&=\sum_{i = 1}^{n}\frac{i}{n}\cdot \frac{1}{n}\\
&=\sum_{i = 1}^{n}\frac{i}{n^2}\\
&= \frac{1}{n^2} \cdot \frac{n(n + 1)}{2}\\
&= \frac{n^{2} + n}{2n^{2}}\\
&= \frac{1}{2} + \frac{1}{2n}\\
\end{align*} 

We can further conclude that as $n$, or the range of the LiDAR sensors increases, 
\[
\lim_{n \to \infty} \frac{1}{2}+\frac{1}{2n} = \frac{1}{2},
\]
though LiDAR sensors usually will not detect objects further than a few hundred meters away.

Taking these values and introducing them into the equation for the probability of danger, the outcome is 
\[
P(D) = \frac{5}{12} \cdot \frac{1}{2} = \frac{5}{24} \approx .2083.
\]

The simulation operates under the same assumptions as the previous calculations, with $n$ set to 60. With $n = 60$, $P(E_1) \approx .5083$, and $P(D)$ is calculated to be $.2118$. To determine the accuracy of this calculation and the previous hypothesis, Algorithm \ref{alg:mc} was formulated.

\begin{algorithm}
    \caption{Monte Carlo simulation for ICanC system} \label{alg:mc}
    \begin{algorithmic}[1]
    \Require{$n$ (Highest number of boxes generated)}
    \Ensure{$Averages$}
    \Statex
    \State $Averages \gets []$ 
    \For{$i = 1$ to $n$}
        \State $sum \gets 0$
        \State $num\_trials \gets 1000$ \Comment{Monte Carlo trials}
        \For{$j = 1$ to $num\_trials$}
            \State $boxes \gets \text{GenerateBoxes}(i)$ \Comment{$i$ boxes}
            \State PublishBoxes($boxes$) \Comment{Pass boxes into system}
            \State $pct\_danger \gets \text{GetPercentDanger}(i)$ \Comment{Determine danger percentage given \textit{i} boxes published}
            \State $sum \gets sum + pct\_danger$ 
        \EndFor
        \State $average \gets sum / num\_trials$
        \State \text{Append} $average$ \text{to} $Averages$
    \EndFor
    \State \Return $Averages$
    \end{algorithmic}
    \label{alg:1}
\end{algorithm}

\begin{figure}
    \centering
    \includegraphics[width=1\linewidth]{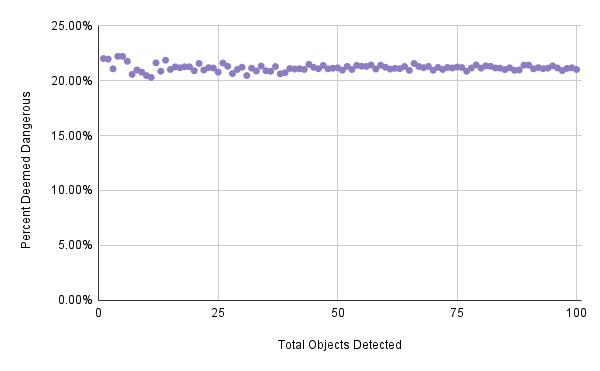}
    \caption{Results from Monte Carlo Simulation on probability that the Danger Detector will label an object dangerous.}
    \label{fig:mcData}
\end{figure}

Running this algorithm outputs the results in Fig \ref{fig:mcData}, which will be discussed further in the evaluation.

It should be noted that this simulation operates under assumptions that are not entirely realistic, as objects in physical environments are rarely random. For instance, the vast majority of vehicles will follow a predictable path, adhering to traffic laws and lane lines. Nonetheless, this simulation offers insight into the predicted behavior of the Danger Detector as well as an estimate of the amount of danger detected by the node.

\section{Evaluation}
\label{sec:5}

The ICanC system was extensively tested in numerous environments, and was further evaluated on multiple metrics: GPU/CPU usage, latency, and energy conservation, along with the standard performance metrics. 

\subsection{Performance Metrics}
Data was recorded during physical testing in the form of rosbag files, a format for recording ROS messages such as point cloud and image data. Each bag encompassed a unique scenario, where a pedestrian would either start close to or far from the vehicle. The pedestrian would then move toward, away from, or perpendicular to the vehicle. Each scenario was predetermined to have objects deemed as truly dangerous or not dangerous. True positives, true negatives, false positives, and false negatives were then determined, and the performance metrics in Table \ref{tab:metrics} were calculated.  

The metric most relevant to ICanC is recall, which weighs true positives and false negatives. ICanC controls a light that assists camera object detection; there are negligible consequences for turning on the light more often than needed. However, failing to detect danger and outputting false negatives pose a substantial threat to the vehicle and its passengers. The recall score of the model is an acceptable score of 72.73\%.

\subsection{System Utilization}
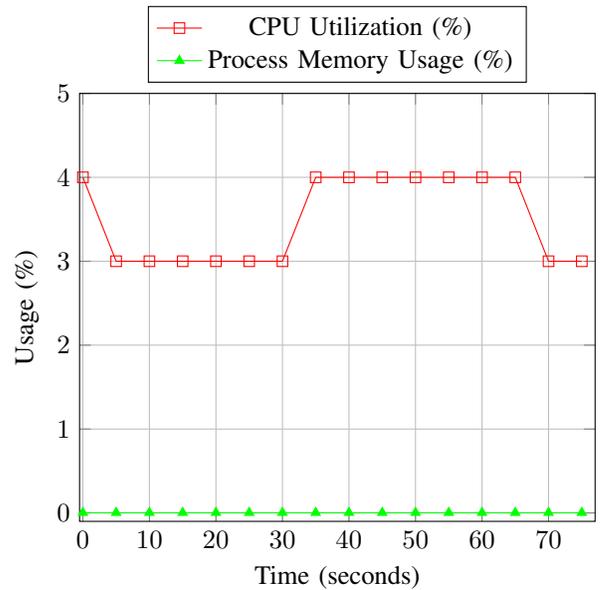
\begin{figure}[t]
    \centering
\begin{tikzpicture}
    \begin{axis}[
        xlabel={Time (seconds)},
        ylabel={Usage (\%)},
        xtick={0,10,...,75},
        ymin=-.1,
        ymax=5,
        xmin=-.5,
        xmax=77,
        legend style={at={(0.5,1.025)}, anchor=south},
        grid=major
    ]

    % CPU Utilization
    \addplot[
        color=red,
        mark=square,
        solid
    ] coordinates {
        (0,4) (5,3) (10,3) (15,3) (20,3) (25,3) (30,3) (35,4) (40,4) (45,4) (50,4) (55,4) (60,4) (65,4) (70,3) (75,3)
    };
    \addlegendentry{CPU Utilization (\%)}

    % Process Memory Usage
    \addplot[
        color=green,
        mark=triangle*,
        solid
    ] coordinates {
        (0,0.00302) (5,0.00302) (10,0.00302) (15,0.00302) (20,0.00302) (25,0.00302) (30,0.00302) (35,0.00302) (40,0.00302) (45,0.00302) (50,0.00302) (55,0.00302) (60,0.00302) (65,0.00302) (70,0.00302) (75,0.00302)
    };
    \addlegendentry{Process Memory Usage (\%)}

    \end{axis}
\end{tikzpicture}

    \caption{CPU usage plotted against time}
    \label{fig:cpu}
\end{figure}

\begin{table}[t]
    \centering
    \caption{Performance Metrics}
    \begin{tabular}{lc}
        \toprule
        Metric & Value\\
        \midrule
        Precision & 57.14\%\\
        Recall & 72.73\% \\
        Accuracy & 74.29\%\\
        F1 Score & 64.00\%\\
        \bottomrule
    \end{tabular}
    \label{tab:metrics}
\end{table}

GPU utilization, CPU utilization, and memory usage were evaluated through the rosbags recorded in the physical environment. The system was run on a computer equipped with a Nvidia GeForce RTX 2080 GPU, and the system utilization was recorded every five seconds. The entirety of ICanC, including sensors and object detection algorithms, proved stressful on processing units; however, it can be reasoned that LiDAR and camera object detection algorithms, if equipped on an autonomous vehicle, will be initiated regardless. Thus, Fig. \ref{fig:cpu} displays the system utilization when ICanC nodes are isolated from detection algorithms. Overall, it can be concluded that the ICanC system doesn't subject the CPU to intense loads, only taking approximately 4\% of the CPU during use and 0.003\% of process memory, or 0.96 MiB. Furthermore, the system doesn't utilize GPU at all, showing that the ICanC system is exceptionally resource-efficient. 

\subsection{Latency}
\begin{figure} 
    \centering
    \pgfplotsset{scaled y ticks=false}
    \begin{tikzpicture}
        \begin{axis}[
            title={Latency Over Time},
            xlabel={Object Detected},
            ylabel={Latency (seconds)},
            grid=both,
            width=.5\textwidth,
            height=0.4\textwidth,
            cycle list name=color list,
            yticklabel={\pgfmathprintnumber[fixed, precision=2]{\tick}}, % Ensure fixed-point notation
            ymin=0, % Set the minimum value for y-axis
            ymax=0.05, % Set the maximum value for y-axis
            xmin=0,
            xmax=100
            ]
            % Data points (first 100 entries)
            \addplot coordinates {
                (1, 0.0282359099992)
                (2, 0.0298605950002)
                (3, 0.028218165)
                (4, 0.027097532)
                (5, 0.0264822749996)
                (6, 0.0259764809998)
                (7, 0.0265618259991)
                (8, 0.0267615039993)
                (9, 0.0246574799994)
                (10, 0.028620395)
                (11, 0.0251293279998)
                (12, 0.025266486)
                (13, 0.0274880799998)
                (14, 0.0253932360001)
                (15, 0.024889194)
                (16, 0.0241698540003)
                (17, 0.0242801720005)
                (18, 0.0234242910001)
                (19, 0.0255732850001)
                (20, 0.0257323699998)
                (21, 0.0243931810001)
                (22, 0.0253138690005)
                (23, 0.0257052029992)
                (24, 0.0225442129995)
                (25, 0.0249196000004)
                (26, 0.0228785039999)
                (27, 0.0248821089999)
                (28, 0.0225165930005)
                (29, 0.0223726169997)
                (30, 0.0235841110007)
                (31, 0.0250540679999)
                (32, 0.0237955109997)
                (33, 0.0262432380005)
                (34, 0.0251357629995)
                (35, 0.0245380260003)
                (36, 0.025085323)
                (37, 0.0252422450003)
                (38, 0.0279939470001)
                (39, 0.027068415)
                (40, 0.0271652900001)
                (41, 0.0271892500004)
                (42, 0.0285457659993)
                (43, 0.0264855140003)
                (44, 0.0282920979998)
                (45, 0.0262552880004)
                (46, 0.0286382349996)
                (47, 0.0296750340003)
                (48, 0.0267845590006)
                (49, 0.0269571480003)
                (50, 0.0278920769997)
                (51, 0.0264975740001)
                (52, 0.0305974889998)
                (53, 0.0282332139996)
                (54, 0.0325685500002)
                (55, 0.0280560970004)
                (56, 0.0271437150004)
                (57, 0.0260966709993)
                (58, 0.0275205709995)
                (59, 0.0257489659998)
                (60, 0.0288635479992)
                (61, 0.028244397)
                (62, 0.0263696470001)
                (63, 0.0346772739995)
                (64, 0.0292848099998)
                (65, 0.0278120769999)
                (66, 0.0281631170001)
                (67, 0.0322209389997)
                (68, 0.0273338029992)
                (69, 0.0282952350008)
                (70, 0.0267065759999)
                (71, 0.0286450269996)
                (72, 0.0274860560003)
                (73, 0.0268727029998)
                (74, 0.0268348989994)
                (75, 0.0285922740004)
                (76, 0.0316966780001)
                (77, 0.0319206140002)
                (78, 0.0306478309994)
                (79, 0.0287211270006)
                (80, 0.0260412339994)
                (81, 0.0263755989999)
                (82, 0.0270581710001)
                (83, 0.0268746600004)
                (84, 0.026399759)
                (85, 0.0290576259995)
                (86, 0.0275522660004)
                (87, 0.028281162)
                (88, 0.0257843259997)
                (89, 0.0278143189998)
                (90, 0.0265673209997)
                (91, 0.0287118460001)
                (92, 0.0254742569996)
                (93, 0.0247989630006)
                (94, 0.0261145159993)
                (95, 0.026866702)
                (96, 0.0255214789995)
                (97, 0.0263100809998)
                (98, 0.0249702760002)
                (99, 0.0278010510001)
                (100, 0.0245256110002)
            };
        \end{axis}
    \end{tikzpicture}
    \caption{Latency over time for 100 objects detected}
    \label{fig:latency}
\end{figure}
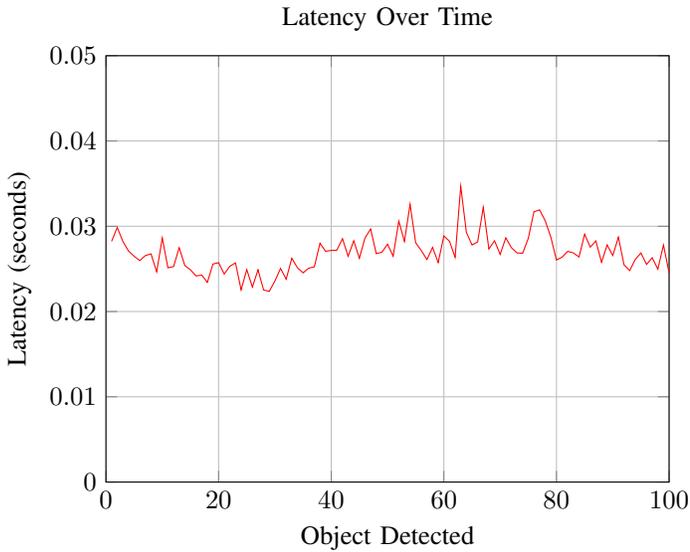

End-to-end latency is recorded through rosbag data from the moment point cloud data is received by the Obstacle Detector to the moment the flashlight receives the command to turn on. The first 100 messages sent were measured and displayed in Fig. \ref{fig:latency}. 

From receiving data to turning on the lights, the latency hovers around 30 ms. This time is well within the bounds of efficient computation given the context of AVs and real-time systems. A latency of 30 ms generally allows AVs or drivers ample time to adjust maneuvering given a dangerous object has been detected. Along with the parameter for reaction time allowed (during testing, three seconds were allowed to react after the light turned on), there should be sufficient time for response.

\subsection{Energy Conservation}
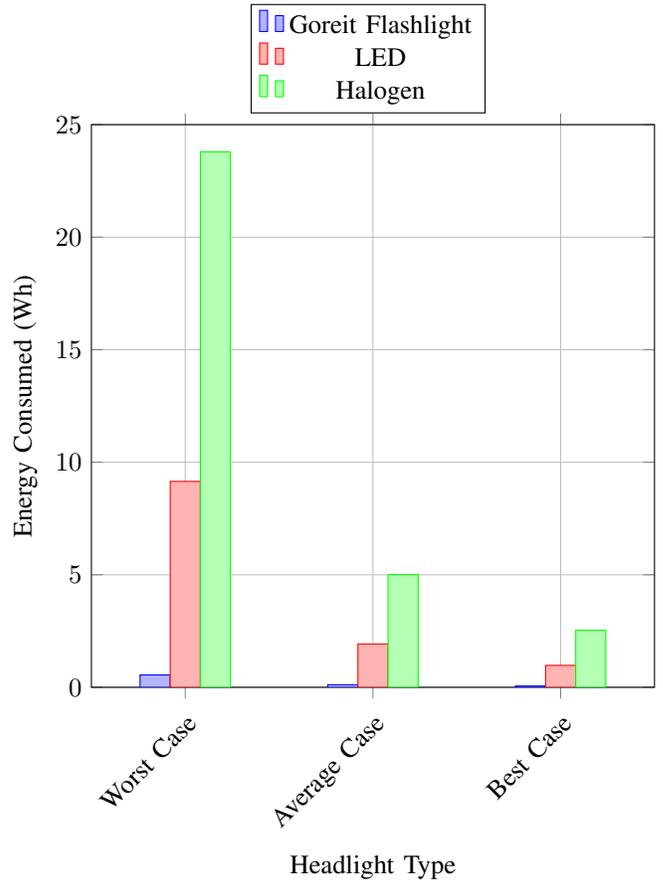
\begin{figure}
\centering
\begin{tikzpicture}
    \begin{axis}[
        width=.5\textwidth, % Scale the plot to \textwidth
        height=0.5\textwidth, % Scale the plot to 0.5\textwidth
        bar width=0.3cm, % Width of bars
        ybar=.05*\pgflinewidth, % Set the bar width
        ymajorgrids = true,
        xlabel = {Headlight Type},
        ylabel = {Energy Consumed (Wh)},
        symbolic x coords = {Worst Case, Average Case, Best Case},
        xtick = {Best Case, Average Case, Worst Case},
        xticklabel style = {rotate=45, anchor=north east},
        legend style = {at={(.7,1.02)}, anchor=south east},
        ymin=0,
        ymax=25,
        enlarge x limits = .25,
        grid = major,
        ylabel = {Energy Consumed (Wh)},
    ]

    % Data for the bar graph
    \addplot[
        ybar,
        color=blue,
        fill=blue!30, % Solid color fill
        bar width=0.4cm
    ]
    coordinates {
        (Best Case, 0.0585)  % Goreit Flashlight energy saved
        (Average Case, 0.1155)
        (Worst Case, 0.549)
    };

    \addplot[
        ybar,
        color=red,
        fill=red!30,
        bar width=0.4cm
    ]
    coordinates {
        (Best Case, 0.975)  % LED energy saved
        (Average Case, 1.925)
        (Worst Case, 9.15)
    };

    \addplot[
        ybar,
        color=green,
        fill=green!30,
        bar width=0.4cm
    ]
    coordinates {
        (Best Case, 2.535)  % Halogen energy saved
        (Average Case, 5.005)
        (Worst Case, 23.79)
    };

    \legend{Goreit Flashlight, LED, Halogen}
    \end{axis}
\end{tikzpicture}
\caption{Energy consumed in the worst case, average case, and best case scenarios for different headlight types.}
\label{fig:energy_saved}
\end{figure}

Energy usage can be measured in terms of wattage $\cdot$ hours of usage. Table \ref{table:wattage} shows the watts used by various relevant lights, including the standard halogen headlights, newer LED headlights, and the Goreit tactical flashlight utilized in the physical testing environment \cite{philips, ultinon}.

To evaluate the potential energy saved for each of these lights, a baseline scenario is considered: a 22-minute drive from the CAR Lab building at the University of Delaware in Newark, DE to Wilmington, DE near midnight. For this scenario, the worst case, average case, and best case for the ICanC system are considered. The system was not run during this drive; instead, energy conserved is calculated for each scenario assuming the system operates precisely how it was designed to operate. The calculations assume usage of high beams, as relatively standard during nighttime (see Fig \ref{fig:energy_saved}).

\subsubsection{Worst Case}
The worst case is defined simply as having headlights on for the entire duration of the drive. This case can be considered the baseline as full usage of headlights is standard in low-illumination environments. By defining the worst case as the baseline, it's ensured that the ICanC system will provide energy-conserving results.

A quick calculation shows that 1.5 watts $\cdot$ 0.366 hours $= 0.549$ Watt-hours (Wh) consumed for the flashlight utilized in the system. For halogen and LED headlights, 23.79 Wh and 9.15 Wh are consumed, respectively. 

\subsubsection{Average Case}
\begin{table}
\centering
\begin{tabular}{| m{5em} | m{5em} | m{5em} |}
\hline
\textbf{Headlight Type} & \textbf{Low Beam (Watts)} & \textbf{High Beam (Watts)} \\ 
\hline
LED & 15 & 25 \\ 
\hline
Halogen & 55 & 65 \\ 
\hline
Goreit Flashlight & N/A & 1.5 \\ 
\hline
\end{tabular}
\caption{Approximate wattage of different headlight types on low and high beam settings}
\label{table:wattage}
\end{table}
Calculations for the probability that a detected object will be deemed dangerous as well as a dedicated Monte Carlo simulation have shown that approximately 21\% of objects will be deemed dangerous (see Fig. \ref{fig:mcData}). Note that, once again, the simulation and the calculations operate under inaccurate assumptions, but assuming that the system detects a random object every second, this data offers an average case benchmark.

The light will be on for 4 minutes and 37 seconds, or .077 hours. Given this, the flashlight, halogen, and LED lights expend 0.1155 Wh, 5.005 Wh, and 1.925 Wh respectively. 

\subsubsection{Best Case}
The ideal case is based on personal observations on the drive from Newark to Wilmington. During the drive, objects believed to be potentially dangerous were counted. These objects included traffic signs and cones close to the vehicle, pedestrians walking on the sidewalk, and other vehicles driving nearby. In the 22-minute drive, 47 objects were personally deemed dangerous. 

Assuming the light timer parameter $\tau$ is set to three seconds, and each object was detected greater than 3 seconds apart, the light should be on for 141 seconds, or .039 hours. The flashlight, halogen, and LED lights would expend 0.0585 Wh, 2.535 Wh, and 0.975 Wh respectively.

Taking a look at each of these scenarios, an average case defined by the Monte Carlo simulation only uses 21\% of the energy that the standard case uses, saving 7.225 Wh for an LED headlight, which AVs are often outfitted with. Furthermore, the best case uses approximately 10\% of the standard case, saving 8.175 Wh for an LED headlight. While the energy savings from minimizing headlight usage may be relatively modest compared to the total energy consumption of the vehicle, it remains a significant consideration. Over time, these incremental savings can accumulate, and any reduction in energy usage is advantageous. 

\section{Discussion}
\label{sec:6}
In this paper, the ICanC system is formulated and implemented. The results suggest efficient processing utilization and speed, acceptable danger-detecting performance metrics, as well as the potential for significant energy conservation improvements in vehicle headlight usage. These findings imply applicability in autonomous vehicles, reducing energy usage without sacrificing accuracy or processing power. 

It's important to note that the focus of this project is on low-illumination states. This project has not been tested in other adverse weather environments that would negatively affect camera-based object detection, including fog, rain, and snow. Most likely, ICanC would perform poorly in these conditions due to LiDAR backscattering. However, as sensor technology continues to evolve, it may become possible to achieve ICanC-esque functionality using weather-independent sensors.

In the future, improvements to ICanC can be achieved by modifying the LiDAR object detection algorithm. Several performance metrics, such as precision and F1 score, were subpar for danger detection. Observations revealed that the object detection algorithm based on Euclidean clustering was quite inaccurate and susceptible to noise. Although the Kalman filter significantly enhanced the accuracy and precision of the tracking algorithm, an improved object detection algorithm would further boost these metrics, one potential avenue being a trained Point Pillars model \cite{lang_2019_pointpillars}.

\section{Conclusion}
\label{sec:7}
In conclusion, ICanC was designed and implemented to promote energy conservation while maintaining accurate camera-based object detection. Extensive testing in real-world and simulated environments demonstrated its computational efficiency, speed, and potential to significantly reduce headlight energy consumption without compromising detection performance. Future work includes generalizing the system to address broader autonomous vehicle challenges and refining the LiDAR detection algorithm. Overall, ICanC successfully achieves its objectives, contributing to the advancement of sustainable transportation.

\section{Acknowledgements}
\label{sec:8}
I would like to express my gratitude to the National Science Foundation for sponsoring this summer research experience at the University of Delaware and for placing me in a position to succeed.

\raggedright
\bibliographystyle{IEEEtran}
\bibliography{icanc}
\fussy

\vspace{12pt}

\end{document}